\pdfoutput=1
\documentclass{bmvc2k}

\usepackage{indentfirst}
\usepackage{graphicx}
\usepackage{color}

\newcommand\blfootnote[1]{%
  \begingroup
  \renewcommand\thefootnote{}\footnote{#1}%
  \addtocounter{footnote}{-1}%
  \endgroup
}

\title{HARRISON: A Benchmark on HAshtag Recommendation for Real-world Images in SOcial Networks}
\addauthor{Minseok Park}{pms0209@kaist.ac.kr}{1}
\addauthor{Hanxiang Li}{hanxiangli@kaist.ac.kr}{1}
\addauthor{Junmo Kim}{junmo@ee.kaist.ac.kr}{1}
\addinstitution{
 School of Electrical Engineering \\
 KAIST \\
 South Korea
}
\runninghead{Park, Li, Kim}{HARRISON: A Benchmark for image hashtag recommendation}

\begin{document}
\maketitle
\begin{abstract}
Simple, short, and compact hashtags cover a wide range of information on social networks. Although many works in the field of natural language processing (NLP) have demonstrated the importance of hashtag recommendation, hashtag recommendation for images has barely been studied. In this paper, we introduce the HARRISON dataset\blfootnote{Dataset is available at \url{https://github.com/minstone/HARRISON-Dataset}}, a benchmark on hashtag recommendation for real world images in social networks. The HARRISON dataset is a realistic dataset, composed of 57,383 photos from Instagram and an average of 4.5 associated hashtags for each photo. To evaluate our dataset, we design a baseline framework consisting of visual feature extractor based on convolutional neural network (CNN) and multi-label classifier based on neural network. Based on this framework, two single feature-based models, object-based and scene-based model, and an integrated model of them are evaluated on the HARRISON dataset. Our dataset shows that hashtag recommendation task requires a wide and contextual understanding of the situation conveyed in the image. As far as we know, this work is the first vision-only attempt at hashtag recommendation for real world images in social networks. We expect this benchmark to accelerate the advancement of hashtag recommendation.
\end{abstract}
\section{Introduction}
\label{sec:intro}
\begin{figure}[t]
\includegraphics[width=\textwidth]{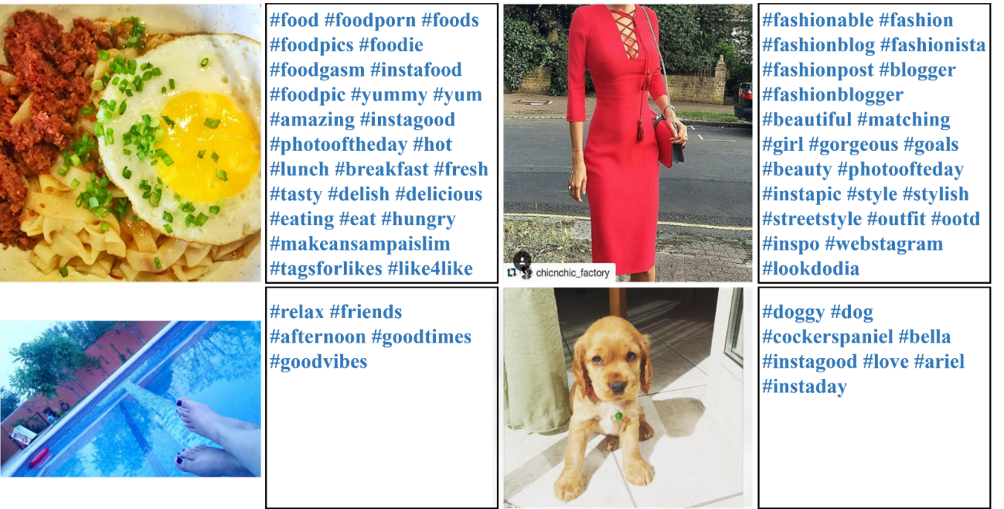}
\caption{Examples of Instagram photos and their hashtags}
\label{ex of insta}
\end{figure}

A hashtag is defined as any word attached to the prefix character \textit{'\#'} that is used in online social network services (SNS) such as Facebook, Twitter, and Instagram. With the growth of online social networks, hashtags are commonly used to summarize the content of a user's post and attract the attention of followers. In Instagram, for example, simple hashtags such as \textit{\#dog} and \textit{\#beach} describe simple objects or locations in a photo. Emotional hashtags such as \textit{\#happy} express a user's feelings, abstract hashtags such as \textit{\#fashion} and \textit{\#spring} categorize topics, and inferential hashtags such as \textit{\#colourful} and \textit{\#busy} represent situational or contextual information. There are even advertising hashtags such as \textit{\#likeforlike}, which are not related to the photo's content. Figure~\ref{ex of insta} demonstrates examples of posted images and hashtags on Instagram. Considering their wide variety, the recommendation of proper hashtags is a highly interesting and useful task in the age of social media. \par
Hashtag recommendation for Twitter text posts has been actively studied in the field of natural language processing (NLP). Previous works focused on the content similarity of tweets~\cite{zangerle2011recommending, li2011twitter} and unsupervised topic modeling with Latent Dirichlet Allocation (LDA)~\cite{tag_generative_krestel2009, tag_generative_godin2013, tomoha_she2014}. Many other approaches~\cite{mazzia2009suggesting, tag_discriminative_Heymann2008, texttag_Khabiri2012} have also been studied and deep learning approaches~\cite{tag_deep_weston2014} have recently been adopted. Although the importance of hashtag recommendation has been proved by many works in NLP task, hashtag recommendation for images has barely been studied in the field of image understanding. Recently, the study of \cite{denton2015user} is presented for the hashtag recommendation systems, but they made use of image data with additional metadata of user information, such as gender and age. As far as we know, our work is the first vision-only attempt to recommend hashtags for real world images in social networks. \par
In the field of computer vision, studies on image understanding and visual analysis have exploded, with various works attempting to challenge such topics as object classification~\cite{resnet2015he,szegedy2015going,simonyan2014very}, object detection~\cite{ren2015faster}, scene classification~\cite{zhou2014learning}, action recognition~\cite{Wang_2015_CVPR}, image captioning~\cite{karpathy2015deep}, and even visual question answering ~\cite{antol2015vqa}. In particularly, image annotation task~\cite{weston2011wsabie, murthy2015automatic} shares similarity with hashtag recommendation task in regard to the diversity of labels. The labels of annotations mostly consist of surface information such as objects in image and locations of image, but hashtags include the inferential words, which require contextual understanding of images, and trendy words in social networks as well as surface information. For these reasons, hashtag recommendation for images is very challenging and is a worthy topic of research in the field of image understanding. \par
In this paper, we introduce the novel benchmark for image hashtag recommendation, titled HARRISON, or the \textbf{HA}shtag \textbf{R}ecommendation for \textbf{R}eal world \textbf{I}mages in \textbf{SO}cial \textbf{N}etworks. The HARRISON dataset is a realistic dataset, which provides real posted images with their associated hashtags in online social network, Instagram. We also construct the baseline models based on convolutional neural networks (CNN), and then experiment and evaluate the performance of our baselines on the HARRISON dataset. Many previous works have been accelerated by the release of benchmark datasets. For example, annual challenges, called ILSVRC~\cite{russakovsky2015imagenet}, ensure that object classification task surpasses human level performance. The HARRISON dataset is expected to accelerate the research and development of hashtag recommendation task. \par
\section{Constructing the HARRISON Dataset}
\label{sec:dataset}
\subsection{Collecting Images and Hashtags}
It is important for hashtag recommendation systems to suggest hashtags that are frequently used in the real world. This means that we need to collect Instagram photos based on the most popular hashtags. A hashtag ranking website\footnote {\url{http://top-hashtags.com/instagram}} is consulted and 50 hashtags are manually selected out of top 100 based on the meaning of hashtags. \par
Next, we collect a set of Instagram photos for each selected hashtag. Through the platform's public APIs (Application Program Interfaces) in a social network analysis website\footnote{\url{https://netlytic.org/index.php}}, photos and their associated hashtags of public posts are collected. For each selected hashtag, list of recent Instagram photos tagged with it and associated hashtags for each photo is gathered. Removing repeated images from the lists, a total of 91,707 images are obtained, with an average of 15.5 associated hashtags per image. The total number of hashtags in this collected dataset is approximately 1.4 million from 228,200 kinds of words. \par
\begin{figure}[t]
\includegraphics[width=\textwidth]{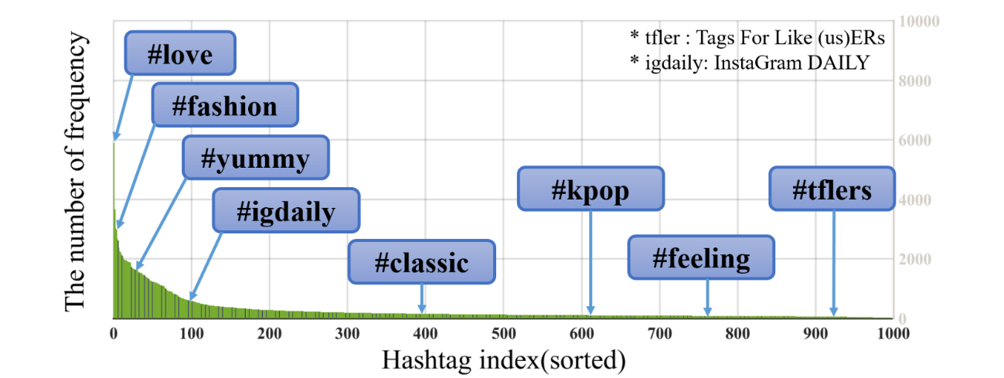}
\caption{Histogram of hashtags in the HARRISON dataset}
\label{fig:statistic}
\end{figure}
\subsection{Organizing the Dataset}
Since there are no rules for making and tagging hashtags, they can be diversely generated and freely used. For example, there are simple cognate hashtags in both the singular and plural form (e.g. \textit{\#girl}, \textit{\#girls}), hashtags in the lower and upper case (e.g. \textit{\#LOVE}, \textit{\#love}), hashtags in various forms of the same root word (e.g. \textit{\#fun}, \textit{\#funny}), sentence-like hashtags (e.g. \textit{\#iwantthissomuch}, \textit{\#kissme}), slang-inspired hashtags (e.g. \textit{\#lol}), and meaningless hashtags to gain the attention of followers (e.g. \#like4like, \#followforfollow). Moreover, users can repeatedly tag the same hashtag for emphasis. \par
To construct a high quality dataset, post-processing on hashtags are performed. First, hashtags containing non-alphabetic characters, e.g. Chinese and Korean, are rejected. Next, lemmatization~\cite{bird2009natural} is applied to all hashtags, which is the process of grouping the different inflect forms of a word. For example, \textit{\#walked}, \textit{\#walks}, and \textit{\#walking} have the same base form \textit{\#walk}. Based on lemmatization process, repeated hashtags associated with the same image are removed. \par
\begin{figure}[t]
\includegraphics[width=\textwidth]{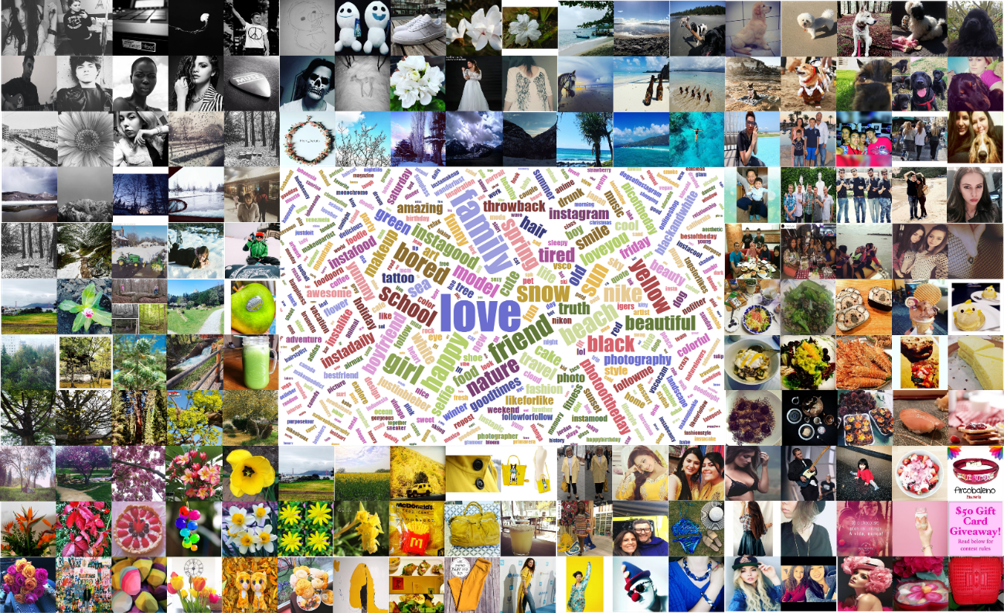}
\caption{ Overview of images and hashtags in the HARRISON dataset }
\label{fig:OverviewHARRISON}
\end{figure}
Through the above processes, approximately 165,000 unique hashtags are collected. Observing the frequency of these hashtags, the top 1,000 hashtags such as \textit{\#love} and \textit{\#friend} form 59\% of the total hashtags and the top 5,000 hashtags account for 74.6\% of the total. Since the rest of hashtags which appear less than 22 times are unused (e.g. \textit{\#mancrushsunday}) or out of style (e.g. \textit{\#sightseeing}), we choose only the 1,000 most frequently used hastags as classes of the dataset and the others are eliminated from the dataset. Finally, we check the number of associated hashtags per image and exclude images with either no hashtag or more than 10 hashtags since photos with too many hashtags are likely to be commercial posts. In this manner, the HARRISON dataset is organized with selected images and corresponding stemmed hashtags from 1,000 classes. \par
\begin{figure}[h]
\includegraphics[width=\textwidth]{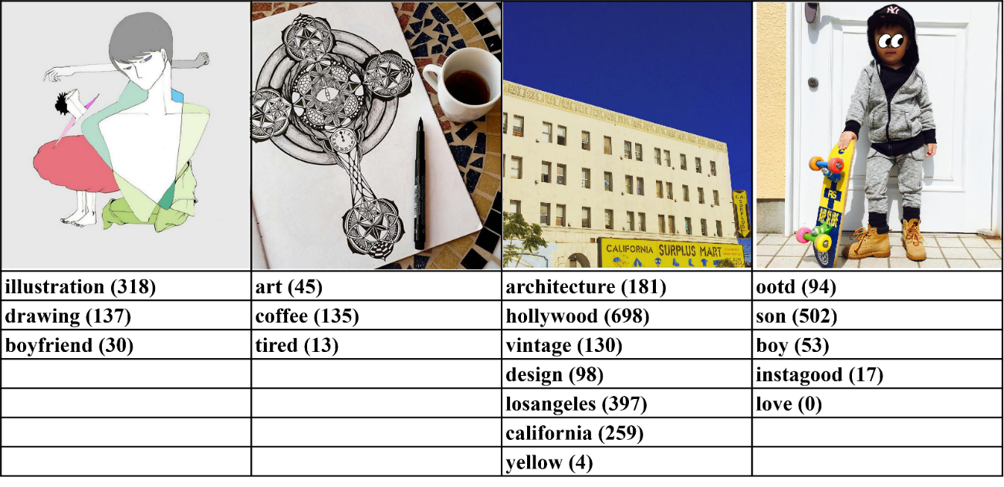}
\caption{Examples of the HARRISON dataset, consisting of images and hashtags with encoded class numbers}
\label{fig:HARRISON}
\end{figure}
\subsection{Description of the HARRISON Dataset}
Figure~\ref{fig:OverviewHARRISON} shows overall distribution of the HARRISON dataset. The HARRISON dataset has a total of 57,383 images and approximately 260,000 hashtags. Each image has an average of 4.5 associated hashtags (minimum 1 and maximum 10 associated hashtags) due to the removal of infrequently used hashtags. The ground truth hashstags for each image are made up of the 1,000 most frequently used hashtags, encoded with numbers based on frequency ranking results, as shown in Figure~\ref{fig:HARRISON}. For evaluation, we randomly split the dataset into 52,383 data for training and 5,000 data for test. \par
\section{Baseline Methods}
\label{sec:baseline}
\begin{figure}[t]
\includegraphics[width=\textwidth]{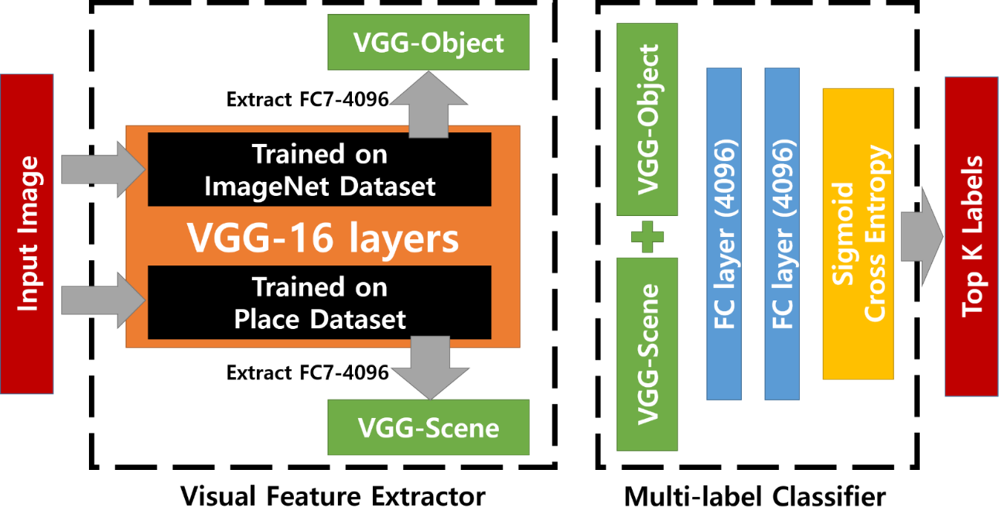}
\caption{Overall flow of baseline methods }
\label{fig:baseline}
\end{figure}

We consider the hashtag recommendation task as a multi-label classification problems. The baseline algorithm consists of two main steps, the visual feature extractor and the multi-label classifier, as shown in Figure~\ref{fig:baseline}. In the visual feature extractor, two kinds of visual features are extracted from the input image. Next, the extracted features are used as the inputs of a trained multi-label classifier, and the score of each class of hashtag is obtained. The scores are sorted, and the top ranked hashtags are recommended for the input images. Details of each step are presented below. \par
In the visual feature extractor stage, high-level feature representation and the diversity of visual information are required. To learn the deep hierarchical features, the VGG-16 layers model~\cite{simonyan2014very} is used as the feature extractor. In addition, we extract two different visual features to obtain the diversity of visual information. Since object and scene task are well-organized on the large-scale dataset, we select object-based and scene-based features as a representative visual features. We train our feature extractor on two different datasets. One is trained on 1.2 million images of the ImageNet dataset~\cite{russakovsky2015imagenet} and the other is trained on 2.5 million images of the Places Database~\cite{zhou2014learning} with 205 scene categories. According to \cite{zhou2014learning}, a simple visualization of the receptive fields for CNN units of two models shows that object-based CNN and scene-based CNN differ in their internal representations. The visual features extracted from the final fully connected layer of each model have 4,096 dimensions, denoted by \textit{VGG-Object} and \textit{VGG-Scene}, respectively. \par
Next, a multi-label classifier with two fully-connected layers and a sigmoid cross entropy layer is constructed. The concatenation of the previous two visual features ($N=4,096 \times 2$) is used as the input of the classifier, and each fully connected layer of classifier have 4,096 neurons. The output of this multi-label classifier is the confidence scores of the 1,000 hashtag classes. To recommend the $K$ number of hashtags, we sort the scores and extract the top $K$ classes. \par
We first construct single feature-based models, which use only \textit{VGG-Object} features or only \textit{VGG-Scene} features as the input of classifier. In these models, unused features are zero-padded in the input layer of classifier. For comparison with these single feature-based models, the integrated model, \textit{VGG-Object} + \textit{VGG-Scene}, is constructed using two visual features. In training phase, the feature extractor and the classifier are trained separately. \par
\section{Experiments}
\label{sec:experiments}
We evaluated the baseline approaches described in Section~\ref{sec:baseline} on our HARRISON dataset. The experiment was performed using the open source Caffe library~\cite{jia2014caffe} on a single NVIDIA GeForce Titan X. In the visual feature extractor, we use pre-trained VGG-16 layers model on ImageNet dataset~\cite{russakovsky2015imagenet} and Place Database~\cite{zhou2014learning}. Using two visual features, we trained our multi-label classifier for 100,000 iterations with 100 samples per mini-batch. The learning rate is initially set to $10^{-1}$, and then is decreased by a factor of 10 every 30,000 iterations. \par
\subsection{Evaluation Measures}
In previous works on tag recommendation~\cite{tag_deep_weston2014, tag_generative_godin2013} and image annotation~\cite{weston2011wsabie, murthy2015automatic}, two performance measures have been mainly used. The first one is precision, which shows how well hashtags are predicted, and the other one is recall how well recommendation system covers the wide range of hashtags. Only precision measure is risky since reasonable hashtags can be missed in real hashtags. In particularly, precision at 1 and recall at 10 is commonly used. The detail explanations of parameters are explained below. In this paper, we select three measures for the fair evaluation of hashtag recommendation, similar to \cite{denton2015user}, Precision and recall are primarily used, and accuracy is also evaluated to check how many mistakes this system makes. \par
For each image, $Precision@K$ is defined as the portion of hashtags in the top $K$ ranked hashtags which match with the ground truth hashtags. $Recall@K$ is defined as the portion of hashtags in the ground truth hashtags which match with the top $K$ ranked hashtags. $Accuracy@K$ is defined as 1 if at least one match between the top $K$ ranked hashtags and the ground truth hashtags exists. Equations of three measures are shown as follows:
\begin{displaymath}
Precision@K = \frac{| Result(K) \cap GT |}{|Result(K)|} 
\end{displaymath}
\begin{displaymath}
Recall@K = \frac{| Result(K) \cap GT |}{|GT|} 
\end{displaymath}
\begin{displaymath}
Accuracy@K = \left\{ \begin{array}{ll} 
1 & \textrm{if} \quad Result(K) \cap GT \neq\emptyset \\  
0 & \textrm{if} \quad Result(K) \cap GT =\emptyset
\end{array} \right.
\end{displaymath}
where $Result(K)$ corresponds to a set of the top $K$ hashtags in predicted results and $GT$ corresponds to a set of the ground truth hashtags. In our experiments, we set $K$ to 1 for precision and 5 for recall and accuracy considering the average number of associated hashtags per image in the HARRISON dataset. \par
\begin{figure}[t]
\includegraphics[width=\textwidth]{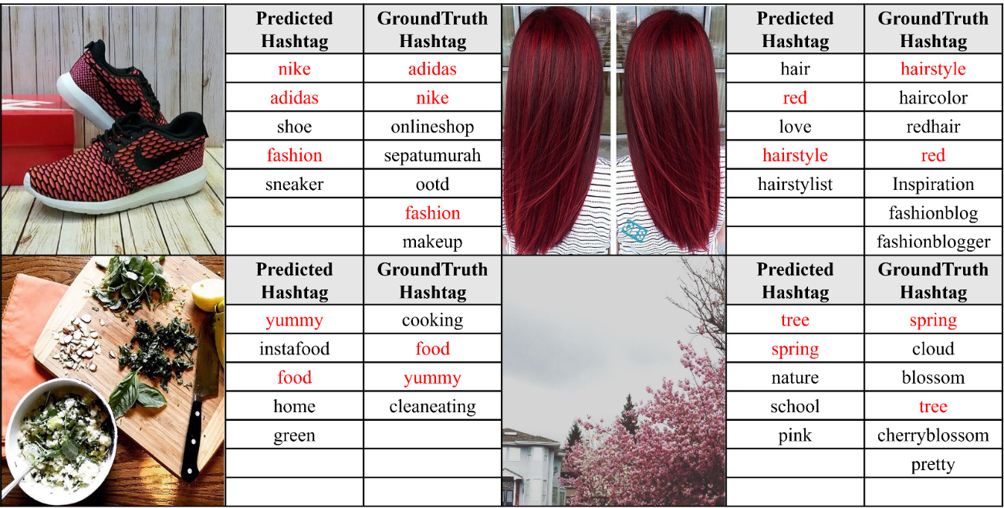}
\caption{Examples of successful hashtag recommendation results with the baseline model \textit{VGG-Object + VGG-Scene}. Matched hashtags between the prediction and ground truth are indicated by red colour.}
\label{fig:success_case}
\end{figure}
\begin{table}[t]
\caption{ Evaluation results of the baseline methods with average precision@1, average recall@5, and average accuracy@5.}
\begin{center}
\begin{tabular}{|l|c|c|c|}
\hline
Baseline Methods 			            			& Precision@1 	& Recall@5 	    	& Accuracy@5 		\\
\hline\hline
\textit{VGG-Object}  	                			& 28.30 \%		& 20.83 \%  		&  50.70 \% 			\\
\textit{VGG-Scene}                     		    	& 25.34 \%		& 18.66 \%  		&  46.30 \% 			\\
\textit{VGG-Object} + \textit{VGG-Scene}            & 30.16 \%    	& 21.38 \%	    	&  52.52 \% 			\\
\hline
\end{tabular}
\end{center}
\label{result_baseline}
\end{table}
\subsection{Results}
We evaluated the baseline results on $Precision@1$, $Recall@5$, and $Accuracy@5$ by averaging over all images in the test set. Table~\ref{result_baseline} shows the evaluation results for our baseline methods on the HARRISON dataset. Compared with \cite{denton2015user}, all three measures in our results were distributed higher, since the number of hashtag classes in the HARRISON dataset is 10 times smaller while the average number of associated hashtags per image is about 2 times greater. \par
\begin{figure}[t]
\includegraphics[width=\textwidth]{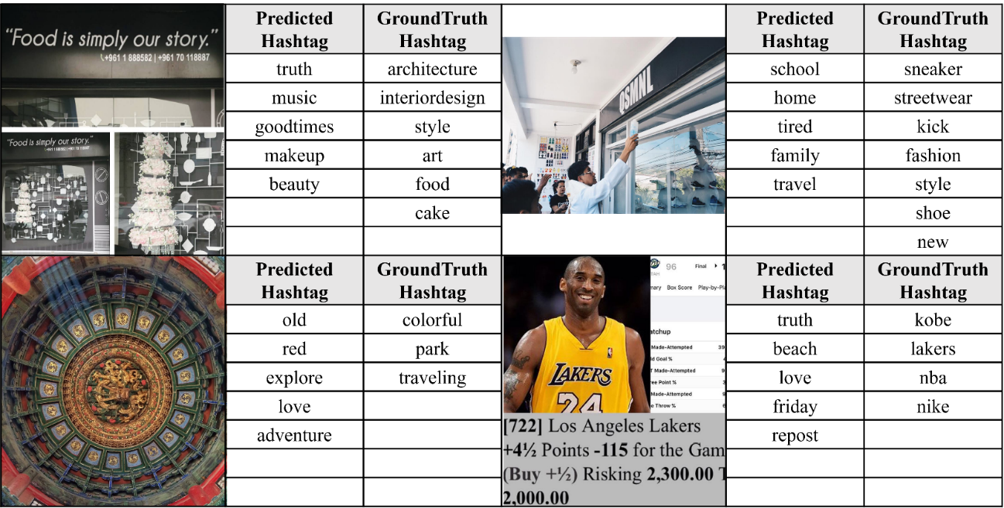}
\caption{Examples of failure cases for the baseline model.}
\label{fig:failure_case}
\end{figure}
\begin{figure}[t]
\includegraphics[width=\textwidth]{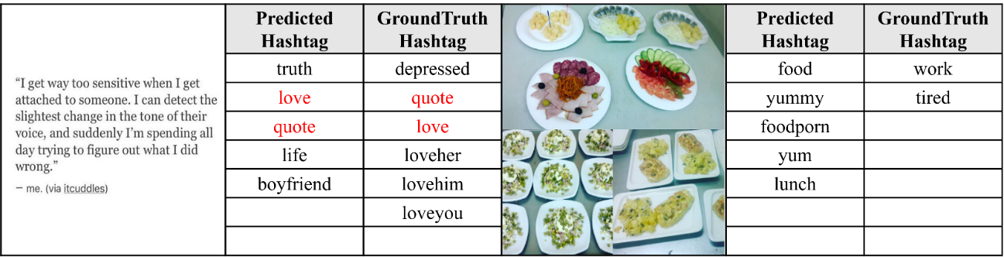}
\caption{Examples of challenging cases and their hashtag recommendation results by the baseline model.}
\label{fig:challenge_case}
\end{figure}
As shown in Table~\ref{result_baseline}, the integrated \textit{VGG-Object} + \textit{VGG-Scene} model achieved the best performance among the baseline models, with the average precision of 30.16 \%, the average recall of 21.38 \%, and the average accuracy of 52.52 \%. Due to the gap in visual information, the results of other models which use a single visual feature were slightly lower. They also show that object-based features are closer to the information available in hashtags than scene-based features. \par
Figure~\ref{fig:success_case} - \ref{fig:challenge_case} demonstrate the examples of baseline results for the HARRISON dataset in various cases. Simple non-inferential labels, such as objects and colours, are well detected by our baseline model. For example, in Figure~\ref{fig:success_case}, \textit{\#hair}, \textit{\#food}, \textit{\#shoe}, \textit{\#red}, \textit{\#green}, \textit{\#pink} are detected. Among these labels, fine-grained classes and non-salient objects are still difficult problems. For example, in Figure~\ref{fig:failure_case}, \textit{\#kobe} is too specific and \textit{\#shoe} is too imperceptible to detect. Even if these problems can be solved by increasing training data, inferential hashtags which are not directly extracted from images are highly challenging problems. For example, we should inference \textit{\#colourful} from the various colours, \textit{\#depressed} from the content of the quoted text in image, and \textit{\#tired} from the many same dishes which user maybe prepared for a long time, as shown in Figure~\ref{fig:challenge_case}. \par
As we mentioned above, success cases of our baseline algorithms are mostly simple non-inferential labels. These observations are supported by the performances of our baseline methods, which are relatively high in $Precision@1$ and $Accuracy@5$ and relatively low in $Recall@5$. This facts show that two visual features are outstanding for the simple non-inferential labels, but insufficient to cover the information available in hashtags. Also, our baseline models ignore the dependencies between hashtags since we considered hashtags as independent labels to use multi-label classifier. Combining with such techniques as word similarity in NLP task could be an option to improve our baseline models. \par
In respect of the HARRISON dataset, our result show that hashtag recommendation task is highly challenging due to the contextual understanding. Further attempt such as multiple instance detection, fine-grained classification, and even text recognition will be helpful to understand contextual information and inference the user's intention. \par
\section{Conclusions}
\label{sec:conclusions}
In this paper, we introduced the HARRISON dataset, a benchmark dataset for hashtag recommendation of real world images in social networks. To evaluate our dataset, we constructed a baseline framework with CNN-based visual feature extractor and multi-label classifier. After applying two visual features, object-based features and scene-based features, we evaluated our baseline models on three evaluation measures. Associated with this dataset, we outlined challenging issues of hashtag recommendation systems: understanding wide range of visual information, using dependencies between hashtag classes, and understanding contextual information. As far as we know, this work presents the first vision-only attempt to recommend hashtags from images. We expect this benchmark dataset to aid the development of hashtag recommendation systems. 
\bibliography{egbib}
\end{document}